\documentclass[letterpaper, 10 pt, conference]{ieeeconf}

\makeatletter
\let\proof\@undefined
\let\endproof\@undefined
\makeatother

\usepackage{cite}
\usepackage{url}
\usepackage{amsmath,amssymb,amsfonts}
\usepackage{algorithmic}
\usepackage{graphicx}
\usepackage{textcomp}
\let\labelindent\relax
\usepackage{enumitem}
\usepackage{multirow}
\usepackage{booktabs}

\usepackage{nicefrac}
\usepackage{subcaption}

\usepackage[ruled,vlined,linesnumbered]{algorithm2e}
\usepackage{algorithmic,float}

\SetKw{Continue}{continue}

\newcommand{\vect}[1]{\boldsymbol{#1}}
\newcommand{\vphi}{\vect{\phi}}

\newcommand{\be}{\begin{equation}}
\newcommand{\ee}{\end{equation}}

\newcommand{\Xcal}{\mathcal{X}}
\newcommand{\Acal}{\mathcal{A}}

\newcommand{\Pp}{\mathbb{P}}

\newcommand{\w}{\vect{w}}

\newcommand{\Pbest}{P^{\mathtt{best}}}
\newcommand{\A}{\vect{A}}

\newcommand{\F}{\mathcal{F}}
\newcommand{\Rcal}{\mathcal{R}}
\newcommand{\wuser}{\w^{\mathtt{user}}}
\newcommand{\Puser}{P^{\mathtt{user}}}

\newcommand{\MEL}{\mathtt{Entropy}}
\newcommand{\MRL}{\mathtt{Regret}}

\usepackage{amsthm}
\theoremstyle{definition}
\newtheorem{problem}{Problem}

\newtheorem{definition}{Definition}

\newtheorem{example}{Example}
\newtheorem*{remark*}{Remark}

\def\BibTeX{{\rm B\kern-.05em{\sc i\kern-.025em b}\kern-.08em
		T\kern-.1667em\lower.7ex\hbox{E}\kern-.125emX}}

\urldef{\smith}\url{stephen.smith@uwaterloo.ca}
\urldef{\wilde}\url{dana.kulic@monash.edu}
\urldef{\kulic}\url{nwilde@uwaterloo.ca}

\title{Active Preference Learning using Maximum Regret}

\author{Nils~Wilde,~
	Dana~Kuli\'{c},~
	and~Stephen~L.~Smith
	\thanks{This research is partially supported by the Natural Sciences and Engineering Research Council of Canada (NSERC).}
	\thanks{N.~Wilde and S.~L.~Smith are with the Department of Electrical and Computer Engineering, University of Waterloo, D.~Kuli\'c is with the University of Waterloo and Monash University. (\wilde; \kulic; \smith)}
}

\begin{document}

\maketitle
\setcounter{secnumdepth}{2}

\begin{abstract}
    We study active preference learning as a framework for intuitively specifying the behaviour of autonomous robots. A user chooses the preferred behaviour from a set of alternatives, from which the robot learns the user's preferences, modeled as a  parameterized cost function. Previous approaches present users with alternatives that minimize the uncertainty over the parameters of the cost function. However, different parameters might lead to the same optimal behaviour; as a consequence the solution space is more structured than the parameter space. We exploit this by proposing a query selection that greedily reduces the maximum error ratio over the solution space. In simulations we demonstrate that the proposed approach outperforms other state of the art techniques in both learning efficiency and ease of queries for the user. Finally, we show that evaluating the learning based on the similarities of solutions instead of the similarities of weights allows for better predictions for different scenarios.
	
\end{abstract}

\section{Introduction}
Recently, research in human robot interaction (HRI) has focused on the design of frameworks that enable inexperienced users to efficiently deploy robots \cite{dragan_orig, sadigh2019, user_study_paper, IRL_apprentice_learning, lfc_manipulator, LfD_trajecotires}.
Autonomous mobile robots for instance are capable of navigating with little to no human guidance; however, user input is required to ensure their behaviour meets the user's expectations. For example, in industrial facilities, a robot might need to be instructed about the context and established workflows or safety regulations \cite{hr_workspace}, or an autonomous car should learn which driving style a passenger would find comfortable \cite{driving_lfd,tunable_traj_planner_urban}.
Users who are not experts in robotics find it challenging to specify robot behaviour that meets their preferences \cite{user_study_paper}. 

\begin{figure}[t]
		\centering
		\begin{subfigure}[b]{0.2\textwidth}
            \centering
            \includegraphics[width=\textwidth]{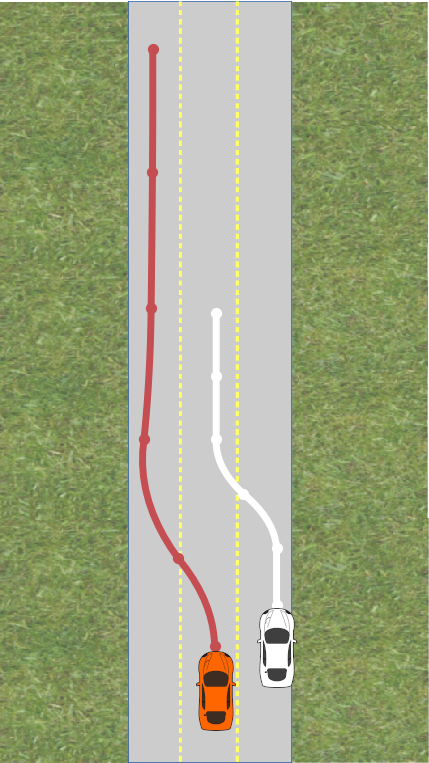}
            \caption{Optimal behaviour.}
        \end{subfigure}%
        \hfill
        \begin{subfigure}[b]{0.2\textwidth}
            \centering
            \includegraphics[width=\textwidth]{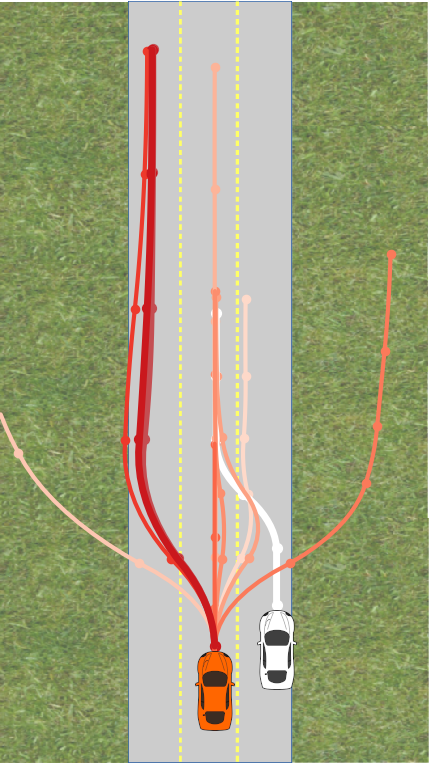}
            \caption{Learned behaviour.}
        \end{subfigure}%
		
		\caption{Behaviour of an autonomous car (red) in the presence of another vehicle (white). In (a) we show the optimal behaviour for some user. In (b) we show alternative paths presented during active preference learning. Darker shades of red indicate behaviour that was presented later. The figure was created using code from \protect\cite{sadigh2019}.}
		\label{fig:intro_example}
\end{figure}

Active preference learning offers a methodology for a robot to learn user preferences through interaction \cite{dragan_orig, sadigh2019, activeRewardLearning, deep_RL_prefs, user_study_paper, RAL_2019}. Users are presented with a sequence of alternative behaviours to a specific robotic task and choose their preferred alternative. 
Figure \ref{fig:intro_example} shows an example of learning user preferences for an autonomous vehicle where alternative behaviours are presented on an interface.
Usually, the user is assumed to make their choice based on an internal, hidden cost function. The objective is to learn this cost function such that a robot can optimize its behaviour accordingly. Often, the user cost function is modelled as a weighted sum of predefined features \cite{dragan_orig}. Hence, learning the cost function is reduced to learning the weights. The key questions in this methodology are  (1) how to select a set of possible solutions that are presented to the user such that the cost function can be learned from few queries to the user, and (2) can the user choose reliably between these solutions.\\

In this work we propose a new approach for selecting solutions in active preference learning.  
In contrast to the work of \cite{dragan_orig, dragan2, sadigh2019} our approach does not focus on reducing the uncertainty of the belief over the weights; instead, we consider the set of all possible solutions to the task. 
Different weights in the user cost function might correspond to similar or even equal optimal solutions; in optimization problems this is known as sensitivity \cite{lin_opti_book}. Thus, even if the estimated weights do not equal the true user weights, the corresponding solution might be the same.
Therefore, we propose a new measure for active preference learning: The regret of the learned path. The concept of regret is known in robust shortest path problems \cite{robust_shortest_path, robust_SP_2approx}. Consider two sets of weights for a user cost function, one that is optimal for a user and one that was estimated through active preference learning. The regret of the estimate captures the suboptimality of the solution found using the estimated weights, i.e., the ratio of the cost of the estimated solution \emph{evaluated} by the optimal weights and the cost of the optimal solution, evaluated by the optimal weights.

We use the notion of regret to select alternatives to show to the user. 
From a set of solutions that are considered equally good for the user given the feedback obtained so far, we choose the pair of solutions $(P, Q)$ that, if $P$ is the optimum, the ratio of costs is maximised. 
As the user either rejects $P$ or $Q$ we remove the most sub-optimal alternative from our solution space.
In each iteration, our proposed approach optimizes over the set of all solutions that are consistent with the user feedback obtained so far. It then presents the user with the pair of solutions where the regret is maximized.

Following the motivation for regret, we evaluate the results of active preference learning based on the learned solution, instead of the learned weights. Therefore, we use the relative error in the cost of paths as a metric. This mirrors how an actual user would evaluate a robot's behaviour: Users are not interested in what weights are used by a robot's motion planner; one of the main motivations for active preference learning is that users find it challenging to express weights for cost or reward functions. Instead users judge a robot's behaviour by how similar it is to what they deem as optimal.

\subsection{Related Work}
The concept of learning a hidden cost or reward function from a user is widely used in various human-robot interaction frameworks, such as learning from demonstrations (LfD) \cite{IRL_apprentice_learning, sadigh_prefdem_2019}, learning from corrections \cite{lfc_uncertainty, dragan_corrections} and learning from preferences \cite{dragan_orig,sadigh_prefdem_2019, dragan2, RAL_2019, user_study_paper}.


Closely related to our work, the authors of \cite{dragan_orig, dragan2} and \cite{sadigh2019} investigate how active preference learning can be used to shape a robot's behaviour. Thereby  a general robot control problem with continuous states and actions is considered. The user cost function is modelled as weighted sum of features. They show that the robot is able to learn user preferred behaviours from few iterations using active preference learning. In \cite{dragan_orig} and \cite{dragan2}, Dragan and colleagues investigate a measure for selecting a new pair of possible solutions to be shown to the user based on the posterior belief over the set of all weights. In detail, new solutions are selected such that the integral over the unnormalized posterior, called volume, is minimized in expectation. This approach is revised in \cite{sadigh2019}, where a failure case for the volume removal is demonstrated. As an alternative measure, the authors propose the information entropy of the posterior belief over the weights.
We show that both of the above approaches disregard the sensitivity of the underlying motion planning problem: Learning about weights of a cost function can be inefficient, as different weights can lead to the same optimal behaviour. In our previous work \cite{RAL_2019} we discretized the weight space into \emph{equivalence regions}, i.e., sets of weights where the optimal solution of the planning problem is the same. 

Another concern during active preference learning is to present alternatives to the user that are easy for them to differentiate which leads to a lower error rate. The authors of \cite{teacher_active_learn} investigate strategies for active learning that consider the flow of the queries to reduce the mental effort of the user and thus decrease the user's error rate. Similarly, \cite{sadigh2019} optimizes for queries that are easy to answer. In our work, we present an active query strategy that features these properties intrinsically: By maximizing the regret of the presented paths, we automatically choose paths that are different with respect to the user cost function and thus are expected to be easily distinguishable for the user.

\subsection{Contributions}
We contribute to the ongoing research in active preference learning as a framework for specifying complex robot behaviours.
We propose a measure for evaluating the solution found by preference learning based on the robot's learned behaviour instead of the learned weights in the cost function.
Further, we propose a new active query selection guided by the maximum error ratio between solutions. Thereby, users are presented with the pair of solutions that has the maximum error ratio among all paths in the feasible solution space.
We demonstrate the performance of our approach by comparing it to a competing state of the art technique and show that our proposed method learns the desired behaviour more efficiently. Moreover, the queries the user is presented with are easier to answer and thus lead to more reliable user feedback.
Finally, we demonstrate how our measure based on solutions gives better predictions about the behaviour of the robot in different scenarios that were not part of the learning.

\section{Problem Statement}

\paragraph{Preliminaries}
Let $\Xcal$ be the state space of a robot and the environment it is acting in and $x_0$ some start state. Further, we have an action space $\Acal$ where each $a\in \Acal$ potentially only affect parts of the state, i.e., there might be static or dynamic obstacles unaffected by the robot's actions.

Further let $P$ be a path of finite length starting at $x_0$.
A path is evaluated by a column vector of predefined features $\vphi(P)=[\phi_1(P), \phi_2(P),\dots, \phi_d(P)]^T$. 
Together with a row vector of weights $\w$ we define the cost of a path as
\be
\begin{aligned}
c(P, \w) =  \vphi(P) \w.
\end{aligned}
\ee
Given some weight $\w'$ let the \emph{optimal path} be $P'=\arg\min_P c(P,\w')$. 
The optimal cost for a weight is 
\be
c^*(\w')=c(P', \w').
\ee
For any other weight $\w$, we call $c(P',\w)$ the cost of $P'$ \emph{evaluated} by $\w$.\\

\paragraph{Problem Formulation}

We consider a robot's state and action space $(\Xcal, \Acal)$ and some start state $x_0$.
Further, let a vector of weights $\wuser$, describing a user's preference for the robot's behaviour and the corresponding optimal path $\Puser$. Each weight $w^{\mathtt{user}}_i$ has a lower and upper bound $l_i$ and $u_i$. However, $\wuser$ itself is hidden. We can learn about $\wuser$ by presenting the user with pairs of paths $(P,Q)$ over $k$ iterations. The objective is to find an estimated path $P^k$ that reflects the user preferences $\wuser$, i.e., is as similar to $\Puser$ as possible.
To evaluate the result of learning $w^{\mathtt{user}}$ the authors of \cite{dragan_orig} propose the \emph{alignment metric}, i.e., the cosine of the angle between the learned weight vector $\w$ and $w^{\mathtt{user}}$.
We adapt this metric and transform it to a normalized error between $0$ and $1$, which we call the \emph{weight error}:
\be
\mathtt{Err}_\mathtt{Weight}(\w, \wuser) = 
\frac{1}{2} 
\left(
1 - \frac{\w\cdot \wuser}{||\w||_2 ||\wuser||_2}
\right).
\label{eq:weight_error}
\ee
The alignment metric was also used in \cite{dragan2, sadigh2019}. However, this metric has two potential shortcomings: 1) It does not consider the sensitivity of the optimization problem that finds an optimal path for a given weight vector. Thus, an error in $\mathtt{Err}_\mathtt{Weight}$ might actually not result in a different optimal path. Moreover, even if the learned weight has a relatively small error, the corresponding path might be suboptimal to the user. 2) The weight error is not suitable as a test error (i.e., to test whether the learned user preferences generalize well to new task instances not encountered during learning) since it does not consider the robot's resulting behaviour: $\mathtt{Err}_\mathtt{Weight}(\w, \wuser)$ is equal for all training and test instances. Hence, the weight error gives no insight into how well the estimated preferences translate into different scenarios, unless $\mathtt{Err}_\mathtt{Weight}(\w, \wuser)=0$, i.e,. the optimal weights are found.
Therefore, we choose a different metric for evaluating the learned behaviour: Instead of the learned weight $\w^k$ we consider the learned path $P^k$. We compare the cost of $P^k$, \emph{evaluated} by the user's true cost $\wuser$ to the optimal cost path of $\wuser$:
\be
\mathtt{Err}_\mathtt{Path}(P^k, \wuser)=
\frac{c(P^k, \wuser)}{c^*(\wuser)} -1.
\label{eq:path_error}
\ee

This error was proposed in \cite{ICRA2018_paper} and we refer to it as the \emph{path error}. 
A similar error was used in \cite{niekum_risk_policy} for finding risk-aware policies in inverse reinforcement learning.
Based on this metric we can now formally pose the learning problem.

\begin{problem}
\label{prob:main_problem}
Given $(\Xcal, \Acal)$ and $x_0$, and a user with hidden weights $\wuser$ who can be queried over $k$ iterations about their preference between two paths $P$ and $Q$, 
find a weight $\w^k$ with the corresponding optimal path $P^k$ starting at $x_0$ that minimizes $\mathtt{Err}_\mathtt{Path}(P, \wuser)$.

\end{problem}

\section{Active Preference Learning}
We introduce the user model and learning framework of our active preference learning approach and then discuss several approaches for selecting new solutions in each iteration.

\subsection{User Model}

To learn about $\wuser$ and thus find $P'$, we can iteratively present the user with a pair of paths $(P,Q)$ and they return the one they prefer:
\be
    \begin{aligned}
        &c(P,\wuser)\leq c(Q,\wuser) \implies \text{ the user returns }P,\\
        &c(P,\wuser)> c(Q,\wuser) \implies \text{ the user returns }Q.
    \end{aligned}
    \label{eq:user_model_det}
\ee
However, a user might not always follow this model exactly. For instance, they might consider features that are not in the model or they are uncertain in their decision when $P$ and $Q$ are relatively similar. 
Thus, we extend equation \eqref{eq:user_model_det} to a probabilistic model, similar to our previous work in \cite{RAL_2019}. Let $I^{P,Q}$ be a binary random variable where $I^{P,Q}=1$ if the user prefers path $P$ over $Q$, and $-1$ otherwise. Then we have 
\be
    \begin{aligned}
        &\Pp\left(I^{P,Q}=1|c(P,\wuser)\leq c(Q,\wuser)\right)
        &=\;& p,\\
        &\Pp\left(I^{P,Q}=-1|c(P,\wuser)\leq c(Q,\wuser)\right)
        &=\;& 1-p,
    \end{aligned}
    \label{eq:user_model_prob}
\ee
where $\nicefrac{1}{2}<p\leq 1$. If $p=1$ we recover the deterministic case from equation \eqref{eq:user_model_det}. In this very simple model the user's choice does not depend on how similar $P$ and $Q$ are.
In the simulations we simulate the user with to the more complex model in \cite{sadigh2019}, which poses the user's error rate as a function of the similarity between alternatives, and show that equation \eqref{eq:user_model_prob} nonetheless allows us to achieve strong performance.

\subsection{Learning Framework}
Over multiple iterations, equation \eqref{eq:user_model_det} yields a collection of inequalities of the form $(\vphi^P-\vphi^Q)\w\leq 0$.
We write the feedback obtained after $k$ iterations as a sequence $U^k=\{(P^1, Q^1),(P^2, Q^2),\dots, (P^{k}, Q^{k})\}$. Without loss of generality, we assume that for any pair $(P,Q)$ in $U^k$ the path $P$ was preferred over the path $Q$.
We then summarize the left-hand-sides $(\vphi^P-\vphi^Q)$ for all $k$ iterations using a matrix $\A^k$. Based on the sequence $U^k$ we can compute an estimate $\w^k$ of $\wuser$ by taking the expectation.

\paragraph{Deterministic case}
In the deterministic case, i.e., $p=1$, the estimate $\w^k$ must satisfy $\A^k\w^k\leq 0$ to be consistent with the user feedback obtained thus far. The set of all such weights constitutes the \emph{feasible set} $\F=\{\w\in \mathbb{R}^{d}| l_i \leq w_i\leq u_i,\;\A^k\w\leq 0 \}$.\\

\subsection{Active Query Selection}

In active preference learning we can choose a pair of paths $(P,Q)$ to present to the user in each iteration $k$.
Throughout this work we only consider paths $P$ and $Q$ that are optimal for some weights $\w^P$ and $\w^Q$.
%
%
Given the user feedback obtained until iteration $k$, a new pair $(P^{k},Q^{k})$ is found by maximizing some measure $f(\w^P,\w^Q, U^{k-1})$ describing the expected learning effect from showing $(P^{k},Q^{k})$ to the user.
Recently, several measures have been introduced:
Removing the \emph{Volume}, i.e., minimizing the integral of the unnormalized posterior over the weights \cite{dragan_orig, dragan2}, maximizing the information entropy over the weights \cite{sadigh2019} and removing \emph{equivalence regions}, i.e., sets of weights where for each weight has the same optimal path~\cite{RAL_2019}.\\

\paragraph{Parameter space and solution space}
The first two approaches maximize information about the parameter space, i.e., the weights $\w$, instead of the solution space, i.e., the set of all possible paths $P$. Despite its motivation based on inverse reinforcement learning, this has a major drawback: The difference in the parameters does not map linearly to the difference in the features of corresponding optimal solutions. Given some $\w^P$ and $\w^Q$, we can compute optimal paths $P$ and $Q$ with features $\vphi^P$ and $\vphi^Q$, respectively. Then $||\w^P-\w^Q||\propto ||\vphi^P-\vphi^Q||$ does \emph{not} necessarily hold.
Thus, learning  efficiently about $\w$ does not guarantee efficient or effective learning about 
paths. Moreover, learning about $\w$ might allow for disregarding a large number of weights.
However, the corresponding optimal paths might be very similar and thus the learning step is potentially less informative in the solution space.
%
\begin{example}
    We consider the autonomous driving example from \cite{sadigh2019} which is posed in a continuous state and action space, illustrated in Figure \ref{fig:intro_example}.
    %
    In Figure \ref{fig:eq_regions_example} we compare the weight error $\mathtt{Err}_{\mathtt{Weight}}(\w, \wuser)$ and the path error $\mathtt{Err}_{\mathtt{Path}}(P, \wuser)$ of $3000$ uniformly random samples. While the weight error is distributed uniformly, the path error distribution takes nearly a discrete form, despite the continuous action space. This illustrates how different weights do not necessarily lead to different solutions, making the solution space more structured than the parameter space.
    \begin{figure} 
    		\centering    		\includegraphics[width=0.49\textwidth]{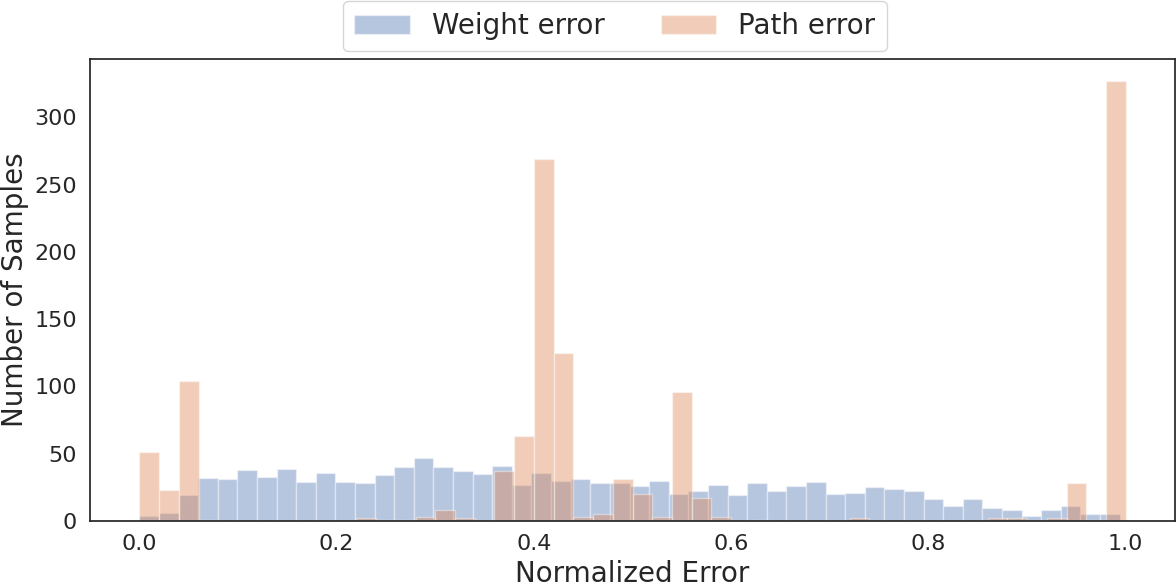}
    		\caption{Example of the sensitivity of a continuous motion planning problem. We show the histogram of the \emph{normalized} weight error $\mathtt{Err}_{\mathtt{Weight}}$ and the \emph{normalized} path error $\mathtt{Err}_{\mathtt{Path}}$ for $3000$ uniformly sampled random weights.}
    		\label{fig:eq_regions_example}
    \end{figure} 
\end{example}
In our previous work \cite{RAL_2019} we proposed a query selection based on a discretization of the weight space: Sets of weights that have the same optimal path are labeled as \emph{equivalence regions}. The objective 
then is to maximally reduce the posterior belief over equivalence regions, i.e, to reject as many equivalence regions as possible. 
A drawback of this approach is that there exists cases where any query only allows for updating the belief of few equivalence regions, resulting in slow convergence.
Because of these limitations of the existing approaches we study a new measure $f(\w^P,\w^Q, U^k)$ based on the solution space.

\section{Min-Max Regret Learning}
\label{Sec:approach}

We propose a new measure $f(\w^P,\w^Q, U^k)$ called the maximum regret, which we seek to minimize.

\begin{definition}[Regret of weights]
    Given a weight $\w^P$ with its corresponding optimal path $P$ and some weight $\w^Q$, the regret of $P$ under $\w^Q$ is 
    \be
       \begin{aligned}
         r(\w^P, \w^Q) 
         &= \frac{c(P, \w^Q)}{c^*(\w^Q)}.
       \end{aligned}
    \ee
\end{definition}
Regret expresses how sub-optimal a path $P$ is when evaluated by some weights $\w^Q$. In active learning, this can be interpreted as follows: \textit{If $P$ is the final estimate, but $Q$ is the optimal solution, how large is the ratio between the cost of $P$, evaluated by $\w^Q$, and the optimal cost?}
We now formulate an approach for selecting which alternatives to show to the user by using regret.
\subsection{Deterministic Regret}
When assuming a deterministic user, we need to assure that $\w^P\in \F^k$ and $\w^Q\in \F^k$, such that the presented paths reflect the user feedback obtained so far. Given $\w^P$ we pose the \emph{Maximum Regret under Constraints Problem (MRuC)} as
\be
    \begin{aligned}
         \max_{\w^Q}&\; r(\w^P,\w^Q) \\
        s.t.&\;  \A^k \w^Q \leq\vect{0},\\
        &\; l_i\leq w^Q_i \leq u_i.
    \end{aligned}
    \label{eq:single_regret}
\ee
The objective can be written in the form $\max_{\w^Q, \vphi^Q} \nicefrac{\vphi^Q\w^Q}{c^*(\w^P)}$. This is a bi-linear program, which are a generalization of quadratic programs. Unfortunately, in our case the objective function is non-convex; generally, such problems are hard to solve.\\

\paragraph{Symmetric Regret}
In equation \eqref{eq:single_regret} we have defined the maximum regret problem when one path is given. While presenting users with a new pair of paths $(P,Q)$, we want to find paths where the regret of $\w^P$ under $\w^Q$ is maximized \emph{and} vice versa.
Thus, we rewrite the objective in \eqref{eq:single_regret} to $r(\w^P,\w^Q)+r(\w^Q, \w^P)$, which we call the \emph{symmetric regret}. 
The maximum symmetric regret of a feasible set $\F^k$ can be found with the following bi-linear program: 
\be
    \begin{aligned}
       \max_{\w^P,\w^Q}&\; r(\w^P,\w^Q) + r(\w^Q,\w^P)\\
        s.t.&\;  \A^k \w^P \leq\vect{0},\\
        &\;\A^k \w^Q \leq\vect{0},\\
        &\;l_i\leq w^P_i \leq u_i,\quad l_i\leq w^Q_i \leq u_i.
    \end{aligned}
    \label{eq:symmetric_regret}
\ee

Similar to equation \eqref{eq:single_regret} this is a non-convex optimization problem. In the evaluation we solve this problem by sampling a set of weights and pre-computing the corresponding optimal paths, following the approach in \cite{dragan_orig}.

\subsection{Probabilistic Regret}
We now formulate regret with consideration of the user's uncertainty when choosing among paths. 
Taking a Bayesian perspective we treat $\wuser$ as a random vector. This allows us to express a posterior belief over $\wuser$ given an observation $I^{P,Q}$. Let $c^P=c(P,\wuser)$ and $c^Q=c(Q,\wuser)$, respectively. Further, we assume a uniform prior over $\wuser$. For any estimate $\w$ where $(\vphi^P-\vphi^Q)\w\leq0$ we have
\be
    \Pp(\wuser=\w|I^{P,Q}) \propto 
    \Pp\left(I^{P,Q}|c^P\leq c^Q\right).
\label{eq:posterior}
\ee
Let $\Pp(\w|I^{P,Q})$ denote $\Pp(\wuser=\w|I^{P,Q}=1)$. We calculate the posterior given a sequence of user feedback $U^k=\{(P^1, Q^1),(P^2, Q^2),\dots, (P^{k-1}, Q^{k-1})\}$ as 
\be
    \Pp(\w|U^k) \propto 
    \prod_{(P,Q)\in U^k}\Pp(\w|I^{P,Q}).
\label{eq:posterior2}
\ee

We formulate the symmetric regret in the probabilistic case by weighting the regret by the posterior of $\w^P$ and $\w^Q$:

\be
   \begin{aligned}
     &\Rcal^k(\w^P, \w^Q | U^k)\\
     &\quad= 
     \Pp(\w^P|U^k) \Pp(\w^Q|U^k)
     \left(
     r(\w^P,\w^Q)+
     r(\w^Q,\w^P)
     \right).
   \end{aligned}
   \label{eq:prob_sym_regret}
\ee

That is, we discount the \emph{symmetric regret} such that we only consider pairs $(P,Q)$ where both $\wuser=\w^P$ and $\wuser=\w^Q$ are likely given the user feedback $U^k$.

Finally, we adapt the problem of finding the maximum symmetric regret from equation \eqref{eq:symmetric_regret} to the probabilistic case. As we cannot formulate a feasible set $\F$ for a probabilistic user, we consider a finite set $\Omega$ where each $\w \in \Omega$ is uniformly randomly sampled from the set $\{\w\in \mathbb{R}^{d}| l_i \leq w_i\leq u_i\}$.
We then take the maximum over all $\w \in \Omega$ to compute the \emph{probabilistic maximum regret}
\be
\Rcal^k_{\max}(U^k) = 
\max_{\w^P, \w^Q\in\Omega} \
    \left[\Rcal^k(\w^P, \w^Q|U^k)\right].
    \label{eq:max_regret}
\ee

In min-max regret learning, we choose the pair of paths $(P,Q)$ that is the maximizer of equation \eqref{eq:max_regret}.

\subsection{Preference Learning with Probabilistic Maximum Regret}
\begin{algorithm}[t]	
	\DontPrintSemicolon 
	\KwIn{$(\Xcal, \Acal),x_0, K$}
	\KwOut{$\Pbest$}		
	Initialize $U^0=\emptyset$\\
	Sample a set of weights $\Omega$\\
	\For{$k=1$ to $K$} {
		
		$\w^P, \w^Q \leftarrow \mathtt{max\_regret}
		(
		(\Xcal, \Acal),x_0, U^{k-1}, \Omega)$\\
		$P,Q\leftarrow \mathtt{opt\_path}(\w^P),\mathtt{opt\_path}(\w^Q)$\\
		$I^k \leftarrow \mathtt{user\_feedback}(P, Q)$\\
		\If{$I^k = -1$}
		{$U^k = U^{k-1} \cup (P,Q)$}
		\Else
		{$U^k = U^{k-1} \cup (Q,P)$}
	
	}	
	\Return{$\mathtt{opt\_path}(\mathbb{E}_{\w}[\w|U^k)])$}   
	   
	\caption{Maximum Regret Learning}
	\label{alg:general}
\end{algorithm}

Our proposed solution for active preference learning using \emph{probabilistic maximum regret} is summarized in Algorithm \ref{alg:general}. In each iteration we find the pair $(\w^p,\w^Q)$ that maximizes the probabilistic symmetric regret as in equation \eqref{eq:max_regret} over a set of samples $\Omega$ (line 4). We then obtain user feedback $I_k=1$ if the user prefers path $P$ and $I_k=-1$ otherwise (line 7) and add the feedback to a sequence (line 6-10). After $K$ iterations, we 
return the path that is optimal for the expected weight, given the observed user feedback (line 11).
Using the maximum regret in the query selection is a greedy approach to minimize the maximum error. Given the current belief over the weights, we choose the pair $P,Q$ with the maximum error ratio, discounted by the likelihoods of $\w^P$ and $\w^Q$.

\section{Evaluation}

We evaluate the proposed approach using the simulation environment from \cite{sadigh2019}, allowing us to compare our approach to theirs in the same experimental setup.
To label the approaches let $\MEL$ denote the maximum entropy learning from \cite{sadigh2019} and $\MRL$ our maximum regret learning.\\

\paragraph{Learning experiments}
First, we consider one of the experiments in \cite{sadigh2019}: The autonomous driving scenario (\emph{Driver}) where an autonomous car moves on a three lane road in the presence of a human-driven vehicle as shown in Figure \ref{fig:intro_example}. Paths are described by four features: Heading relative to the road, staying in the lane, vehicle speed, and the distance to the other car. Every feature is averaged over the entire path. 
Furthermore, we introduce the \emph{Extended Driver} experiment with additional features to create a more complex scenario. In addition to the above features we add the distance travelled along the road, the summed lateral movement, summed and maximum lateral and angular acceleration, the minimal speed, and the minimum distance to the other vehicle.
We choose the Driver example because the entropy approach from \cite{sadigh2019} showed strong results and this scenario was already previously investigated in \cite{dragan_orig}. 
The extension aims to show how the learning techniques behave in higher dimensions. 

Additionally we consider a third experiment adapted from \cite{RAL_2019, user_study_paper}: An autonomous mobile robot navigates between given start and goal locations in a known environment. However, there are $n$ areas in the environment that a user marked as desired or undesired for robot traffic. Each such area is a soft constraint, i.e., there is a penalty or reward associated with it, which can be expressed by a weight. By defining features for all areas describing if a robot trajectory passes through yields a cost function of the form $c(P)=\vphi(P)\w$. Here $\vphi$ is an $n+1$ dimensional vector, the first $n$ entries are the features describing the length of path $P$ in area $i$ for all $i=1,2,\dots, n$. The $n+1$-th feature is the time it takes the robot to execute path $P$.
The robot is unaware of the value of each penalty and reward, i.e., the weights $\w$ are not given to the robot, yielding an instance of Problem \ref{prob:main_problem}. We will refer to this experiment as \emph{Mobile}. The instance of the problem used for evaluation consists of $18$ areas; thus, the dimensionality of the feature and weight space is $19$.

\paragraph{Optimal paths}

Given a weight $\w$ we need to find the corresponding optimal path $P$ in order to evaluate the path error (and to compute regret in Algorithm \ref{alg:general}). In \cite{sadigh2019} no motion planner is given; in the experiments we rely on the generic non-linear optimizer \texttt{L-BFGS} \cite{LBFGS_solver} for the \emph{Driver} and \emph{Extended Driver} experiments. However, depending on the problem, this solver can return suboptimal solutions. To mitigate this effect, we pre-sample paths which are used as a look-up-table. Given a path that was found using \texttt{L-BFGS}, we iterate over all pre-sampled paths; if a sampled path yields a better cost for the given weight, we use that path instead.
In both experiments, the entropy approach uses the implementation provided by \cite{sadigh2019} where queries are chosen from $500,000$ pre-sampled pairs of random paths. Since regret requires optimal paths, the regret approach uses a set of $200$ weights with their corresponding optimal paths, yielding $40,000$ possible pairs of paths%
\footnote{Using sampled optimal paths for the entropy approach did not lead to different results in the experiments, therefore we show the results using the original implementation.}.
Finally, in these experiments the behaviour is actually captured by a reward and not a cost. Thus, an optimal path is found by minimizing the negative cost and we change the definition of regret to $r(\w^P, \w^Q)= 1-\nicefrac{c(P, \w^Q)}{c^*(\w^Q)}$.

In the \emph{Mobile} experiment the robot moves using a state lattice planner \cite{pivtoraiko2009differentially}; given a weight $\w$ we can always find an optimal path in polynomial time. 
We varied the problem setup by choosing three different start and goal locations for the robot to navigate between. For each start goal pair we need to pre-sample paths individually. The discrete state space led to a significantly smaller set of pre-samples, varying between $5$ and $32$.
However, using randomly generated paths as in \cite{sadigh2019} for $\MEL$ led to very poor performance. Therefore, we slightly modified the $\MEL$ approach for this experiment, such that the same pre-samples were used as for the $\MRL$ approach.

\paragraph{Simulated users}
We simulate user feedback using the probabilistic user model from \cite{sadigh2019}. Given two paths, the user's uncertainty depends on how similar the paths are with respect to the cost function evaluated for $\wuser$:
\be
    \Pp\left(I=1|(P,Q),\wuser\right)
    = \frac{e^{c(P,\wuser)}}
    {e^{c(P,\wuser)}+e^{c(Q,\wuser)}}.
    \label{eq:sadigh_user}
\ee

The probabilistic regret is computed using pre-sampled weights as described in Algorithm \ref{alg:general}, with an uncertainty of $p=0.85$ in equation \eqref{eq:user_model_prob}.
Similar to \cite{sadigh2019} for each experiment we sample a user preference $\wuser$ uniformly randomly from the unit circle, i.e., $||\wuser||_2 =1$. We notice that this can include \textit{irrational} user behaviour: A negative weight on heading for instance would encourage the autonomous car to \emph{not} follow the road.



\subsection{Learning error}
\begin{figure}[t]
		\centering
		\begin{subfigure}[b]{0.49\textwidth}
            \centering
            \includegraphics[width=\textwidth]{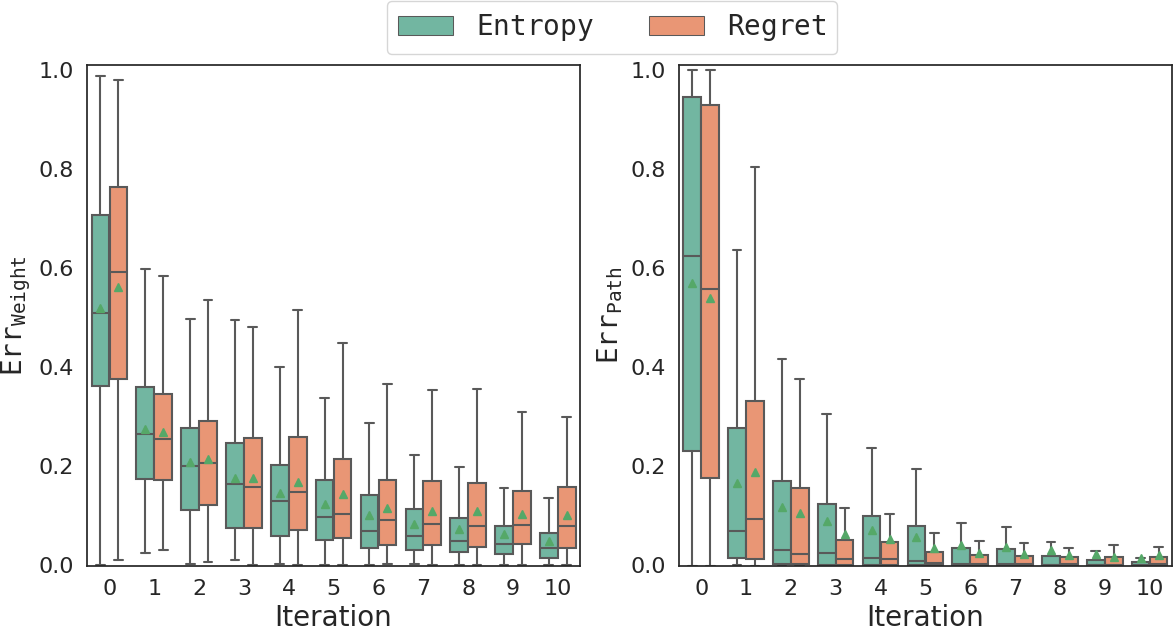}
            \caption{Driver}
            \label{fig:driver_box}
        \end{subfigure}%
        \hfill
        \begin{subfigure}[b]{0.49\textwidth}
            \centering
            \includegraphics[width=\textwidth]{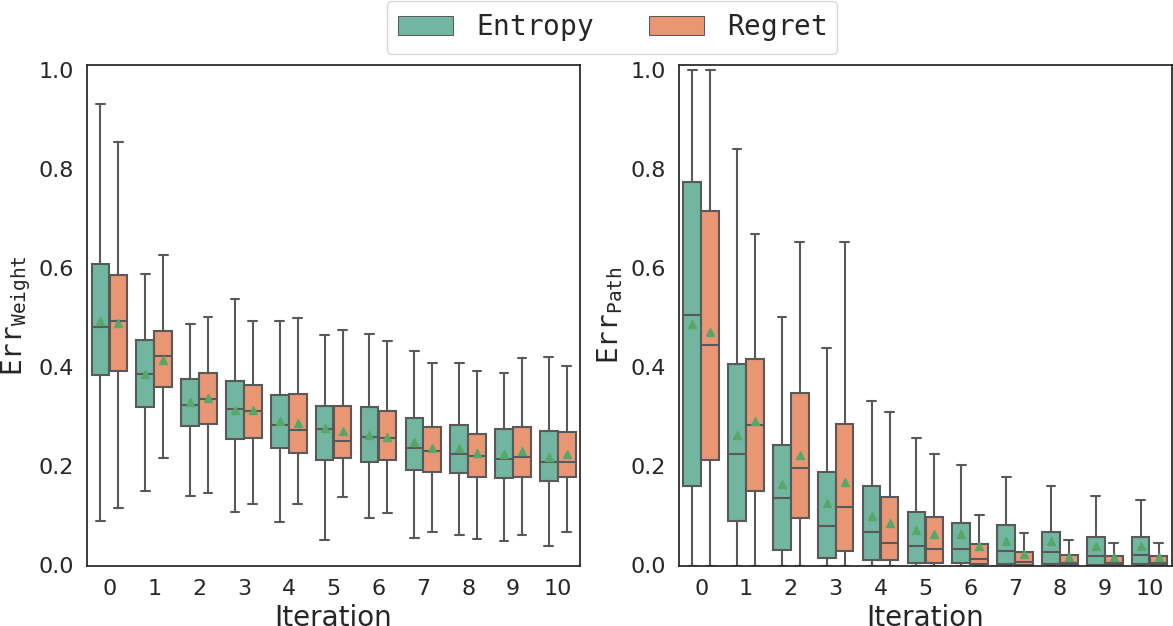}
            \caption{Extended Driver}
            \label{fig:driverExt_box}
        \end{subfigure}%
        \hfill
        \begin{subfigure}[b]{0.49\textwidth}
            \centering
            \includegraphics[width=\textwidth]{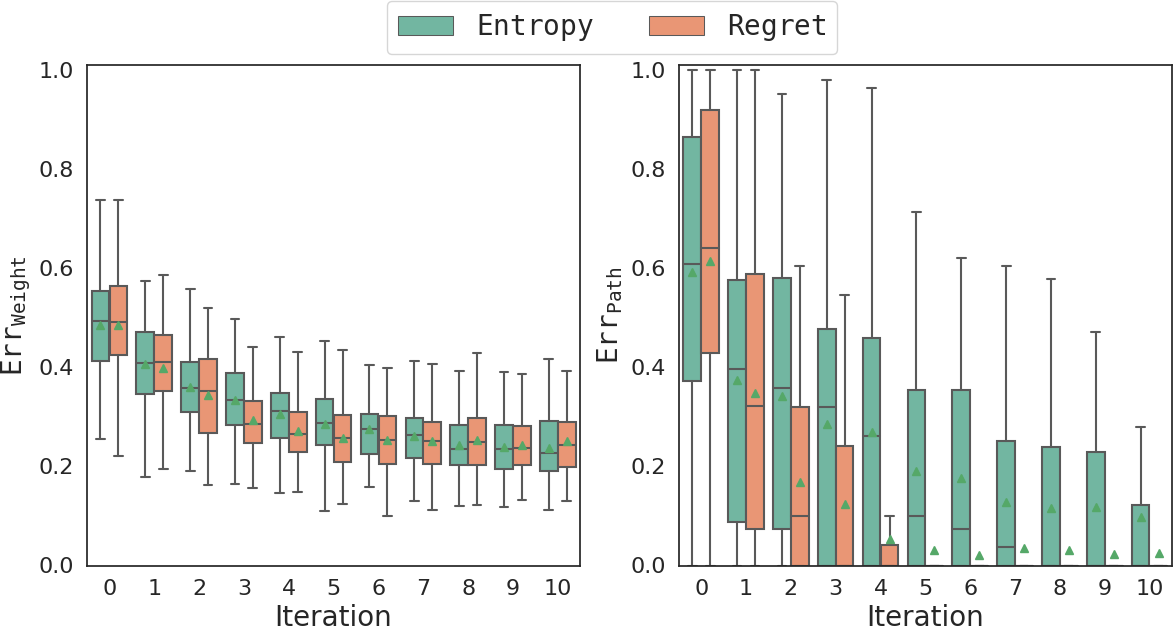}
            \caption{Mobile robot}
            \label{fig:spec_box}
        \end{subfigure}%
		\caption{Comparison of active preference learning with maximizing entropy and minimizing regret.
		}
		\label{fig:experiment2}
\end{figure}

In Figure \ref{fig:experiment2} we compare $\MEL$ to $\MRL$ on both metrics over $10$ iterations for the two experiments, each repeated $200$ times. In the boxplots the center line shows the median and the green triangle shows the mean.


In the driver example, $\MEL$ overall achieves a smaller weight error and smaller deviations from the mean, reproducing the results from \cite{sadigh2019}. In the path space we observe that $\MEL$ achieves a slightly better result in the last two iterations. However, between iteration $2$ and $8$, the $\MRL$ approach performs better, i.e., learns more quickly.
Overall, both approaches perform equally well. 

For the \emph{Extended Driver} example in Figure \ref{fig:driverExt_box}, both approaches make limited progress on the weight metric and exhibit large deviations.
For the path error we observe that $\MEL$ performs better initially, but makes little progress after iteration $6$. The final median lies at $\approx0.05$ at iteration $10$, but the highest quartile still reaches up to $0.2$. The $\MRL$ approach achieves a lower mean and median error in iteration $4$ and subsequently improves further. At iteration $8$ the mean and median are close to $0$, the box plot also shows that three quarters of all trials are very close to convergence.

Figure \ref{fig:spec_box} illustrates the result for the \emph{Mobile} experiment. 
Here, the weight error shows no difference between the two approaches; both perform equally poorly and inconsistently. At the same time the path error shows a large difference. $\MRL$ achieves convergence for nearly all trials after just $5$ iterations (some outliers cause the mean value to still be at $\approx 0.05$). At the same time the performance of $\MEL$ is inferior: Even though the median error becomes $0$ in iteration $8$, the mean value is still $\approx0.15$ with large deviations.

In conclusion, $\MRL$ achieves an equally good result as $\MEL$ on the path error for the \emph{Driver} experiment, despite having a larger weight error. That is, while the weights found by $\MEL$ are more similar to $\wuser$ based on the alignment metric, the resulting behaviour of $\MRL$ and $\MEL$ are equally good. 
Moreover, the \emph{Extended Driver} and \emph{Mobile} experiments have shown that the performance of $\MEL$ deteriorates for higher dimensions, i.e., larger sets of features. In contrast, $\MRL$ still achieves a very strong performance on the path error.
%

\subsection{Easiness of queries}
\begin{figure}[t]
	\centering
	\includegraphics[width=0.49\textwidth]{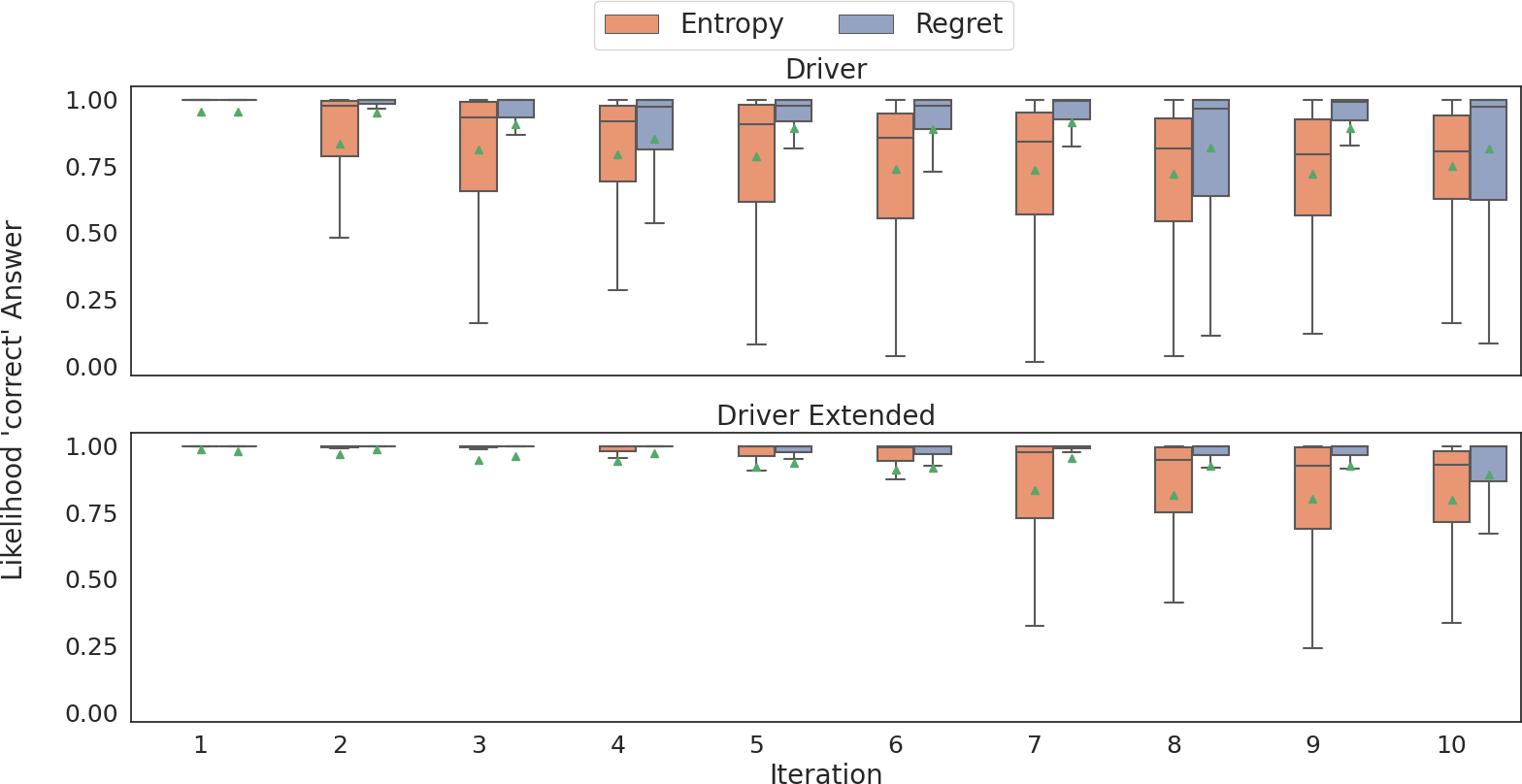}
	\caption{The likelihood that the simulated user gives the 'correct' answer, i.e., the probability in equation \eqref{eq:sadigh_user}.}
	\label{fig:user_error}
\end{figure} 

A major contribution of \cite{sadigh2019} is the design of queries that are \emph{easy} for the user to answer, i.e., the probability that the user choice is inconsistent with the assumed cost function from equation \eqref{eq:sadigh_user} is low. In maximum regret learning we do not directly consider the user's uncertainty when choosing a new pair of paths. However, as the paths maximizing the probabilistic symmetric regret 
have a large difference on cost,
our approach implicitly selects paths $P, Q$ that potentially are easy for a user to answer.
To compare the easiness of the queries presented to the user, we consider the probability that the user would choose the path with lower cost, evaluated by $\wuser$ \eqref{eq:sadigh_user}. 
In Figure \ref{fig:user_error} we compare the probability of correct user answers for $\MEL$ and $\MRL$.

In the \emph{Driver} experiment, we recorded a mean probability of $0.84$ for $\MEL$, which is slightly worse than reported for the strict queries in \cite{sadigh2019}, where correct answers occurred in $87\%$ of cases. 
Nonetheless, with $\MRL$ the simulated answers had a mean probability of being correct of $0.94$, outperforming $\MEL$.
In Figure \ref{fig:user_error}, we observe that both approaches achieve very high probabilities for correct user answers in the first iteration, i.e., ask an \emph{easy} question. Afterwards, the probabilities get smaller: The median of $\MEL$ decreases to $0.9$ in iteration $3$ and the deviations increase significantly. The $\MRL$ approach maintains higher median values for all iterations. 
Interestingly, we observe cyclic decreases of the mean (and increases for the deviations) in iterations $4$, $8$ and $10$. According to the user model the presented paths were very similar, indicating that the learning might have been close to convergence. This aligns with the small errors of the expected weight reported in Figure \ref{fig:driver_box}.

In the \emph{Extended Driver} experiment the user behaviour is much more accurate for both approaches with a mean of $0.92$ for $\MEL$ and $0.96$ for $\MRL$. This indicates that the sampled paths differ more in cost. From iteration $7$ onwards, $\MEL$ starts to show larger deviations, i.e., questions become more difficult to answer, implying that the presented paths are very similar.
Together with the very small decrease in path error the we observed in Experiment 1 (Figure \ref{fig:driverExt_box}) at the same iteration, this leads to the conjecture that $\MEL$ is converging to a local optimum.
Finally, the \emph{Mobile} experiment did not show any difference between the two approaches, both achieving a very high accuracy of $0.99$.

Overall, these results strongly support our claim that maximizing regret implicitly creates queries that are easy for the user to answer.

\subsection{Generalization of the error}
\label{sec:test_error}
Finally, we investigate how the two error metrics generalize to different scenarios, independent of whether the error is a result of learning with $\MEL$ or $\MRL$. That is, we investigate how useful each error metric is for predicting the robot's performance when deployed in a new instance of the problem not encountered during learning. For the \emph{Driver} experiment we use the setup from Figure \ref{fig:intro_example} as a training case and construct five test cases by changing the initial state of the human driven vehicle (white).
The weight error is scenario independent, it directly describes how similar the estimated weight is to $\wuser$. Thus, the weight error is the same in training and test cases and cannot be used as test error, as this would contain no additional information about performance on the test case. Hence, we use the path error as the test error.
Further, we notice that if the weight error is zero, i.e., the weights have been learned perfectly, then the path error is zero in all scenarios. However, as shown in Figure \ref{fig:experiment2} and in \cite{dragan_orig, sadigh2019} the weights typically do not converge to the true user weight within few iterations. 
Given some weight the path errors are fixed values in every test scenario. We are interested in how well the weight error and the path error of the training predict the path error of the test scenario.

\begin{figure} 
	\centering
	\includegraphics[width=0.49\textwidth]{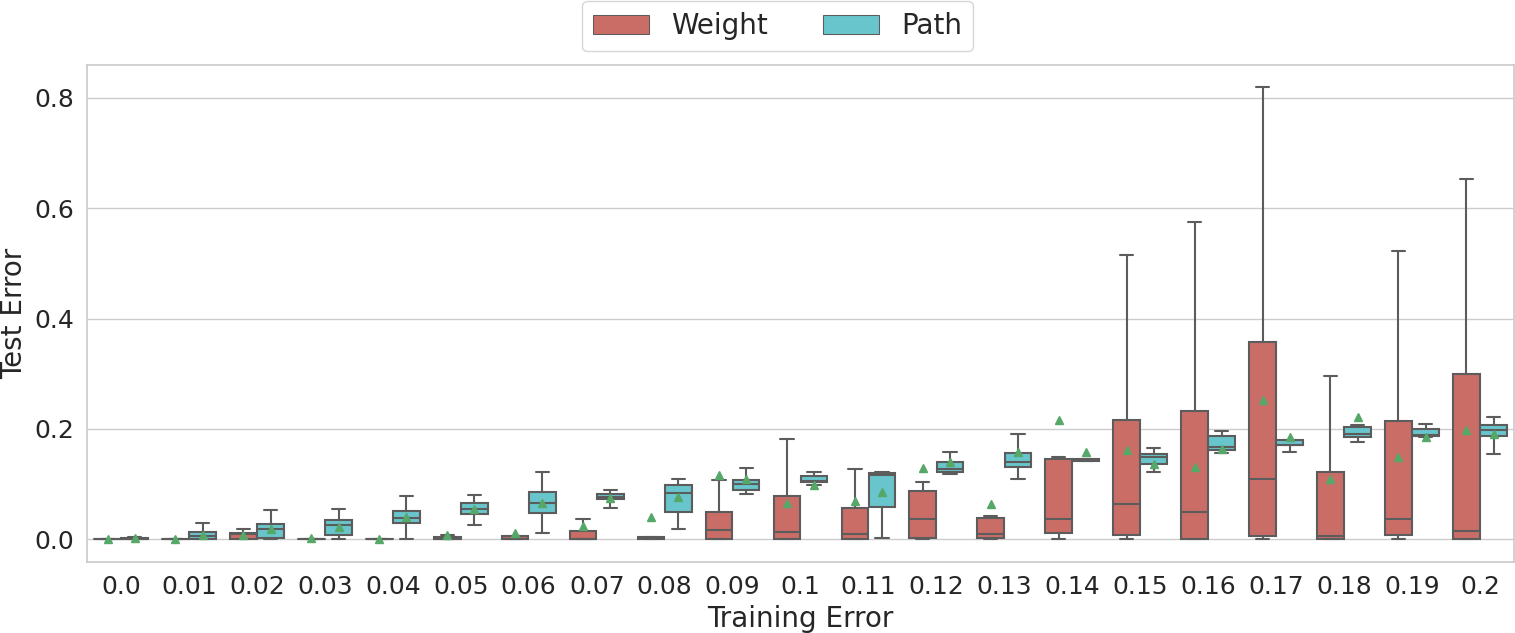}
	\caption{Relationship between training errors measured by the path and weight metric to test errors in the path metric.}
	\label{fig:test_error}
\end{figure}

We generate $40$ different random user weights $\wuser$ and then generate $200$ estimates of each of these weights. For every estimate we find the optimal path and compute the path and weight error which are used as training errors for the estimate. %
In Figure \ref{fig:test_error} we show how these training errors relate to the test error. We compare the path and weight error as a measure of generalisation performance (i.e., how well the weight and path errors predict the test case performance).

We observe that the path error translates linearly between training and test scenarios: Given a weight with a certain path error in the training scenario, the weight yields paths in the test scenarios that have a similar path error, on average.
The relationship between weight error and test error is more complex. For a weight error of $0-0.01$ during training, we observe a test error of $0$, i.e., if the weights are very close to the optimum, the optimal solution is found in every scenario. However, for larger training errors the test error shows large deviations, implying that a low weight error in training is not a robust measure of how good the resulting behaviour is in test cases. The observation is supported by a strong Pearson correlation of $p=0.92$ between training and test error for the path error, but a much weaker correlation of $p=0.28$ for the weight error.
This lends support to the claim that the path error is better suited for making predictions of the performance in scenarios that were not part of the training.

We conducted a the same experiment for the \emph{Extended Driver} scenario. The correlation of the path error is weaker, but with $p=0.80$ still substantially stronger than for the weight error where we observed $p=0.27$.
In the \emph{Mobile} scenario, training and test instances are defined by multiple start-goal pairs. However, we observed no correlation for both path error and weight error. The features in this scenario are local, i.e., describe if the robot visits a certain part of the environment. 
Learning about one task gains insufficient information to always find a good path for a different task.

In summary, we observe that the path error is more suitable than the weight error for reliable predictions of the test performance in scenarios with global features. However, higher dimensions can weaken the reliability, and local features may not allow for any predictions.


\section{Discussion}
In this paper we investigated a new technique for generating queries in active preference learning for robot tasks. We have shown that competing state of the art techniques have shortcomings as they focus on the weight space only. As an alternative, we introduced the regret of the cost of paths as a heuristic for the query selection, which allows to greedily minimize the maximum error. Further, we studied an error function that captures the cost ratio between the behaviour of estimated preferences and the optimal behaviour, instead of the similarity of weights. 
In simulations we demonstrated that using regret in the query selection leads to faster convergence than entropy while the queries are even easier for the user to answer.
Moreover, we have shown that the path error allows for better predictions for other scenarios.

For future work special cases such as discrete action spaces in the form of lattice planners should be investigated. This would give further inside into the computational hardness of finding the maximum regret and potentially allow for solution strategies that do not require pre-sampling weights and paths. Richer user feedback such as an \emph{equal preference} option could be of interest, promising results for this approach were presented in \cite{dragan2, sadigh2019}. Finally, regret based preference learning should be investigated in a user study to show the practicality of this approach.

\bibliographystyle{IEEEtran}

\end{document}